\documentclass[preprint,12pt]{elsarticle}
\usepackage{amssymb}
\usepackage{rotating}
\usepackage{booktabs}
\usepackage{graphicx}
\usepackage{tabularx}
\usepackage{array}
\usepackage{lscape}
\usepackage{longtable}
\usepackage{amssymb}
\usepackage{tabularx}
\usepackage{booktabs}
\usepackage{subfigure}
\usepackage{amsmath}
\usepackage{epsfig}
\usepackage{longtable}
\usepackage{lscape}
\usepackage{epsf}
\usepackage{multirow}

\newdefinition{definition}{Definition}
\newdefinition{example}{Example}
\newdefinition{property}{Property}

\journal{The Scientific World Journal}
\begin{document}
\begin{frontmatter}


\title{D numbers theory: a generalization of Dempster-Shafer evidence theory}


%\author[address1]{Shiyan Huang}
%\author[address2]{Xiaoyan Su}
%\author[address3]{Yong Hu}
%\author[address4]{Sankaran Mahadevan}
\author[address1,address4]{Yong Deng \corref{label1}}

\cortext[label1]{Corresponding author: Yong Deng, School of Computer and Information Science, Southwest University, Chongqing, 400715, China. \\ E-mail address: professordeng@163.com, ydeng@swu.edu.cn}

\address[address1]{School of Computer and Information Science, Southwest University, Chongqing 400715, China}
%\address[address2]{School of Electronics and Information Technology, Shanghai Jiao Tong University, Shanghai 200240, China}
%\address[address3]{Institute of Business Intelligence and Knowledge Discovery, Guangdong University of Foreign Studies, Guangzhou 510006, China}
\address[address4]{School of Engineering, Vanderbilt University, Nashville, TN 37235, USA}

\begin{abstract}
Efficient modeling of uncertain information in real world is still an open issue. Dempster-Shafer evidence theory is one of the most commonly used methods. However, the Dempster-Shafer evidence theory has the assumption that the hypothesis in the framework of discernment is exclusive of each other. This condition can be violated in real applications, especially in linguistic decision making since the linguistic variables are not exclusive of each others essentially. In this paper, a new theory, called as D numbers theory (DNT), is systematically developed to address this issue. The combination rule of two D numbers is presented. An coefficient is defined to measure the exclusive degree among the hypotheses in the framework of discernment. The combination rule of two D numbers is presented. If the exclusive coefficient is one which means that the hypothesis in the framework of discernment is exclusive of each other totally, the D combination is degenerated as the classical Dempster combination rule. Finally, a linguistic variables transformation of D numbers is presented to make a decision. A numerical example on linguistic evidential decision making is used to illustrate the efficiency of the proposed D numbers theory.
\end{abstract}
\end{frontmatter}
% \linenumbers
\section{Introduction}
\label{}

Quantitative handling incomplete, uncertain and imprecise information data warrants the use of soft computing methods \cite{zadeh1984review}. Soft computing methods such as fuzzy set theory \cite{zadeh1965fuzzy}, rough set \cite{pawlak2007rudiments,pawlak2007rough}, Dempster-Shafer evidence theory \cite{dempster1967upper,shafer1976mathematical} can essentially provide rational solutions for complex real-world problems. The traditional Bayesian (subjectivist) probability approach cannot differentiate between aleatory and epistemic uncertainties and is unable to handle non-specific, ambiguous and conflicting information without making strong assumptions. These limitations can be partially addressed by the application of Dempster-Shafer evidence theory, which was found to be flexible enough to combine the rigor of probability theory with the flexibility of rule-based systems \cite{sadiq2006estimating,huang2013new}. Due to its efficiency to handle uncertain information, evidence theory is widely used in many applications such as pattern recognition\cite{denuxdenoeux2006classification}, evidential reasoning \cite{yang2002evidential,yang1994evidential}, complex network and systems \cite{kang2012evidential,wei2013identifying}, DS/AHP \cite{beynon2002ds,beynon2000dempster,beynon2005method,yao2010improved,ma2013model,ju2012emergency,utkin2012combining,deng2011new,beynon2006role,deng2014supplier} and other decision making fields \cite{jousselme2012distances,deng2013topper,liu2012new,wang2013spatial,zargar2012dempster,yang2011risk,denoeux2013maximum}.

However, there are some limitations in the classical Dempster-Shafer evidence theory. One of the well known problems is the conflict management when evidence highly conflicts, which is heavily studied. However, some other issues are paid little attention. For example, the elements in the frame of discernment must be mutually exclusive which has greatly limited its practical application \cite{deng2014environmental,deng2012d}. For example, it is not correct to have a basic probability assignment as $m\{good\}=0.8,m\{good,very good\}=0.2$, since the linguistic variable $very  good$ is not exclusive of the other linguistic variable $good$.

Recently, some applications of D numbers to represent uncertain information has been reported, which is an extension of Dempster-Shafer evidence theory \cite{deng2014environmental,deng2012d,Deng2014DAHPSupplier,Deng2014BridgeDNs} . D numbers can effectively represent uncertain information since that the exclusive property of the elements in the frame of discernment is not required, and the completeness constraint is released if necessary. Due to the propositions of applications in the real word could not be strictly mutually exclusive, these two improvements are greatly beneficial. To get a more accurate uncertain data fusion, a discounting of D numbers based on the exclusive degree is necessary. However, some key issues on D numbers are not yet well addressed. For example, it is necessary to develop a reasonable combination rule of D numbers. In addition, similar to the pignistic probability transformation in belief function theory, the transformation of D numbers into linguistic variables to make a final decision is inevitable. To address these issues, the D numbers theory (DNT) is systematically developed in this paper.

The rest of the paper is organized as follows. Section 2, some preliminaries are briefly introduced. The D numbers theory is presented in Section 3. The application in linguistic decision making is used to illustrate the efficiency of the proposed DNT. Conclusions are given in Section 5.

\section{Preliminaries}
\subsection{Fuzzy set theory}
Fuzzy set theory is widely used in uncertain modelling \cite{deng2012fuzzy,zhang2013ifsjsp}. In some decision makings, assessments are given by natural language in the qualitative form. These linguistic variables can be assessed by means of linguistic terms \cite{zadeh1975concept1,zadeh1975concept2,zadeh1975concept3} which is proposed by Zadeh.
\begin{definition}
Let $X$ be a universe of discourse. Where $\widetilde A$ is a fuzzy subset of $X$; and for all $x \in X$ there is a number ${\mu _{\widetilde A}}(x) \in [0,1]$ which is assigned to represent the membership of $x$ in $A$, and is called the membership of $\widetilde A$. \cite{zadeh1965fuzzy}.
\end{definition}
\begin{definition}
A fuzzy set $\widetilde A$ of the universe if discourse $X$ is convex if and only if for all $x_1$, $x_2$ in $X$,
\begin{equation}
{\mu _{\widetilde A}}(\lambda {x_1} + (1 - \lambda ){x_2}) \ge \min ({\mu _{\widetilde A}}({x_1}),{\mu _{\widetilde A}}({x_2}))
\end{equation}
where $\lambda  \in [0,1]$.
\end{definition}
\begin{definition}
A triangular fuzzy number $\widetilde A$ can be defined by a triplet(a, b, c), as shown in Fig. \ref{membership}.
\begin{equation}
{\mu _{\widetilde A}}(x) = \left\{ {\begin{array}{*{20}{c}}
{0,}&{x < a}\\
{\frac{{x - a}}{{b - a}},}&{a \le x \le b}\\
{\frac{{c - x}}{{c - b}},}&{b \le x \le c}\\
{0,}&{x > c}
\end{array}} \right.
\end{equation}
\begin{figure}[htbp]
\begin{center}
\centerline{
\includegraphics[width=6.5cm]{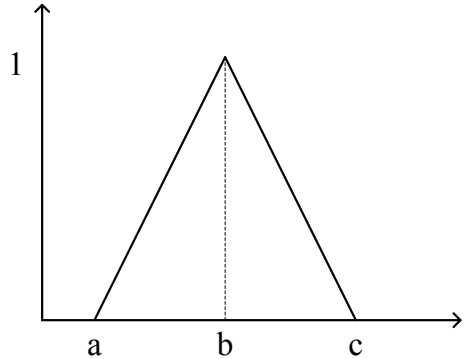}
}
\caption{The membership function.
}\label{membership}
%(d) the US power-grid. (f)network of e-mail interchanges.
\vspace{0.2in}
\end{center}
\end{figure}
\end{definition}
\subsubsection{Linguistic variable}
The concept of a linguistic variable is very useful in dealing with situations which are too complex or poorly-defined to be reasonably described in conventional quantitative expressions. Linguistic variables are represented in words, sentences or artificial languages, where each linguistic value can modeled by a fuzzy set \cite{zimmermann1991fuzzy}. In this paper, the importance weights of various criteria and the ratings of qualitative criteria are considered as linguistic variables. These linguistic variables can be expressed in positive triangular fuzzy numbers, as shown in Tables \ref{VL} and \ref{VP}.
\begin{table}[!h]
{\footnotesize%
\caption{Linguistic variables for the importance weight of each criterion}\label{VL}
\begin{center}
\begin{tabular}{p{4cm}p{2cm}}
\toprule
Very low (VL)    & (0,0,0.1)   \\
Low (L)          & (0,0.1,0.3)   \\
Medium low (ML)  & (0.1,0.3,0.5)   \\
Medium (M)       & (0.3,0.5,0.7)   \\
Medium high (MH) & (0.5,0.7,0.9)   \\
High (H)         & (0.7,0.9,1.0)   \\
Very High (VH)   & (0.9,1.0,1.0)   \\
\bottomrule
\end{tabular}
\end{center}}
\end{table}

\begin{table}[!h]
{\footnotesize%
\caption{Linguistic variables for the importance weight of each criterion}\label{VP}
\begin{center}
\begin{tabular}{p{4cm}p{2cm}}
\toprule
Very poor (VP)    & (0,0,1)   \\
Poor (P)          & (0,1,3)   \\
Medium poor (MP)  & (1,3,5)   \\
Fair (F)          & (3,5,7)   \\
Medium goog (MG)  & (5,7,9)   \\
Good (G)          & (7,9,10)   \\
Very good (VG)    & (9,10,10)   \\
\bottomrule
\end{tabular}
\end{center}}
\end{table}

\subsection{Dempster-Shafer (DS) theory of evidence}
Let $\Theta$ denote a finite nonempty set of mutually
exclusice and exhaustive hypotheses, called \emph{the frame of
decernment}.
\begin{definition}\label{def1}
A mass function is a mapping m: $2^{\Theta}\to[0,1]$, which
satisfies:
\begin{eqnarray}\label{eqt1}
m(\emptyset ) = 0 \quad and \quad \sum\limits_{A \subseteq \Theta} {m(A) = 1}.
\end{eqnarray}
\end{definition}
A mass function is also called a \emph{basic probability
assignment(BPA)} to all subsets of $\Theta$.

\begin{definition}\label{defbp}
The belief Bel(A) and plausibility Pl(A) measures of an event A $\subseteq$ $\Theta$ can be defined as
\begin{eqnarray}\label{eqbp}
Bel(A) = \sum\limits_{A_i: A_i \subseteq A}{m(A_i)}, Pl(A) = \sum\limits_{A_i: A_i \cap A\ne \emptyset}{m(A_i)}.
\end{eqnarray}
\end{definition}

The belief Bel(A) and Plausibility Pl(A) measures can be regarded as lower and upper bounds for the probability of A according to \cite{Dempster1967,halpern1992two}. Let $p_i$ be some unknown probability of the i-th element of the $\Theta$, then the probability distribution p = $(p_1,p_2,$...,$p_m)$ satisfies the following inequalities for all focal elements A:
\begin{center}
Bel(A) $\le$ $\sum\limits_{i: u_i \in A}{p_i}$ $\le$ Pl(A).
\end{center}

\begin{definition}\label{def2}
Discounting Evidences: If a source of evidence provides a mass
function $m$ which has probability $\alpha$ of reliability.
Then the discounted belief m$^\prime$ on $\Theta$ is defined as:
\begin{eqnarray}\label{eq5}
m'(A) = \alpha m(A),\quad \forall A \subset \Theta ,A \ne \Theta.
\end{eqnarray}
\begin{eqnarray}\label{eq6}
m'(\Theta ) = 1 - \alpha  + \alpha m(\Theta ).
\end{eqnarray}
All mass function is discounted by $\alpha$, which is called
discount coefficient.
\end{definition}
\begin{definition}\label{Dempsterrule}
Dempster's rule of combination, denoted by ${(m_1 \oplus m_2)}
($also called the orthogonal sum of $m_1$ and $m_2)$, is defined
as follows:
\begin{eqnarray}\label{Dempsterruleeq}
m(A) = \frac{1}{{1 - k}}\sum\limits_{B \cap C = A} {m_1 (B)m_2
(C)}.
\end{eqnarray}
where
\begin{eqnarray}\label{classicalconflict}
k = \sum\limits_{B \cap C = \emptyset } {m_1 (B)m_2 (C)}.
\end{eqnarray}
\end{definition}

Note that $k$ is called the normalization constant of the
orthogonal sum ${(m_1 \oplus m_2)}$. The coefficient $k$ is also
called as the conflict coefficient between $m_1$ and $m_2$,
denoted as $\mathop m\nolimits_\oplus (\emptyset )$ in the
following of the paper.

\subsection{Pignistic probability transformation(PPT) \cite{smets1994transferable}}
\begin{definition}\label{def4}
Beliefs manifest themselves at two levels - the credal level(from credibility) where belief is entertained, and the pignistic level where beliefs are used to make decisions. The term "pignistic" was proposed by Smets \cite{smets1994transferable} and originates from the word pignus, meaning 'bet' in Latin. Pignistic probability is used for decision-making and uses Principle of Insufficient Reason to derive from BPA. It has been increasingly used in recent years \cite{chen2013fuzzy,du2012method,diaz2012octree}. It represents a point estimate in a belief interval and can be determined as:
\begin{eqnarray}\label{eqbet}
BetP(B) = \sum\limits_{A \in \Theta } {m(A)} \frac{{\left| {B \cap
A} \right|}}{{\left| A \right|}}.
\end{eqnarray}
where $\left| A \right|$ denotes the number of elements of
$\Theta$ in $A$.
\end{definition}

\section{D numbers theory}

In this section, the D numbers theory is developed systematically. There are three main parts of this theory, namely the uncertainty modelling, the combination of D numbers and the decision making based on linguistic variable transformation. These parts are detailed as follows.

\subsection{Definition of D numbers}

In the mathematical framework of Dempster-Shafer theory, the basic probability assignment (BPA) defined on the frame of discernment is used to express the uncertainty in judgement. A problem domain indicated by a finite non-empty set $\Omega$ of mutually exclusive and exhaustive hypotheses is called a frame of discernment. Let $2^\Omega$ denote the power set of $\Omega$, a BPA is a mapping $m: 2^\Omega \to [0,1]$, satisfying
\begin{eqnarray}
m(\emptyset ) = 0 \quad and \quad \sum\limits_{A \in 2^\Omega }
{m(A) = 1}
\end{eqnarray}

BPA has an advantage of directly expressing the ``uncertainty" by assigning the basic probability number to a subset composed of $N$ objects, rather than to an individual object. Despite this, however, there exists some strong hypotheses and hard constraints on the frame of discernment and BPA, which limit the representation capability of Dempster-Shafer theory regarding the uncertain information. On the one hand, the frame of discernment must be a mutually exclusive and collectively exhaustive set, i.e., the elements in the frame of discernment are required to be mutually exclusive. This hypothesis however is difficult to be satisfied  in many situations. For example some assessments are often expressed by natural language or qualitative ratings such as ``Good", ``Fair", ``Bad". Due to these assessments base on human judgment, they inevitably contain intersections. Therefore, the exclusiveness hypothesis cannot be guaranteed precisely so that the application of Dempster-Shafer theory is questionable and limited. On the other hand, a normal BPA must be subjected to the completeness constraint, which means that the sum of all focal
elements in a BPA must equal to 1. But in many cases, the assessment is only on the basis of partial information so that an incomplete BPA is obtained. Fox example in an open world
\cite{smets1994transferable}, the incompleteness of the frame of discernment may lead to the incompleteness of information. Additionally, the Dempster's rule of combination also cannot handle the incomplete BPAs.

D numbers \cite{deng2012d} is a new representation of uncertain information, which is an extension of Dempster-Shafer theory. It overcomes these existing deficiencies in Dempster-Shafer theory and appears to be more effective in representing various types of uncertainty. D numbers are defined as follows.
\begin{definition}
Let $\Omega$ be a finite nonempty set, a D number is a mapping formulated by
\begin{equation}
D: \Omega \to [0,1]
\end{equation}
with
\begin{eqnarray}
\sum\limits_{B \subseteq \Omega } {D(B) \le 1}
\end{eqnarray}
where $B$ is a subset of $\Omega$.
\end{definition}

Compared with the definition of BPA in evidence theory, there are two main differences listed below.
One, the sum of the D numbers is not necessary to 1. The main reason is that only in the close world can we guarantee that $D(\emptyset ) = 0$. The other, $\emptyset$ is an empty set and the following condition is not necessary in D numbers theory since we may be in the open world.
\begin{equation}
D(\emptyset ) = 0
\end{equation}

However, we focus on the situation in close world in this paper. As a result, if we do not specialize to point out, we assume $D(\emptyset ) = 0$ which means the close world. We focus on the handling the non-exclusive hypotheses based on D numbers, especially in linguistic environment.

\begin{figure}[!htbp]
  \centering
  \subfigure[Frame of discernment in Dempster-Shafer evidence theory]{
    \label{FrameDS}
    \includegraphics[scale=0.60,angle=-90]{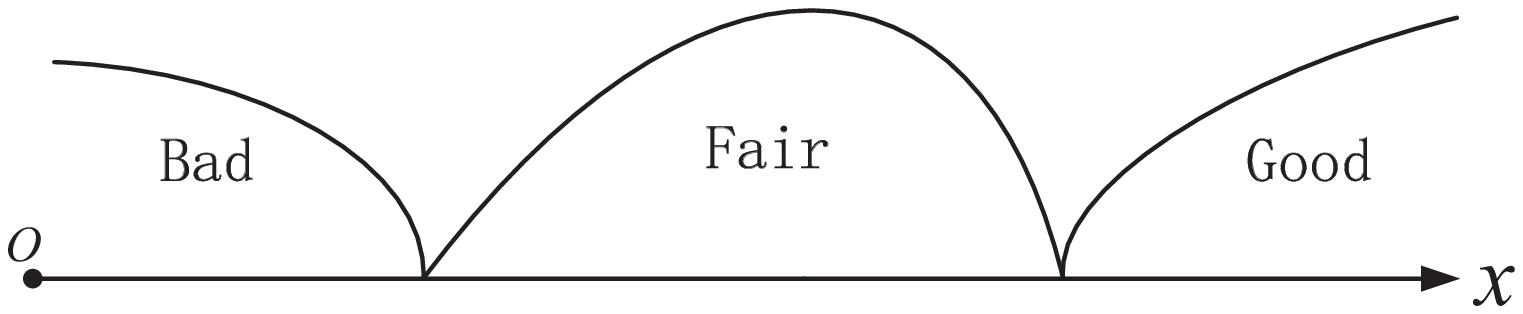}}
  \hspace{0.1in}
  \subfigure[Problem domain in D numbers]{
    \includegraphics[scale=0.60,angle=-90]{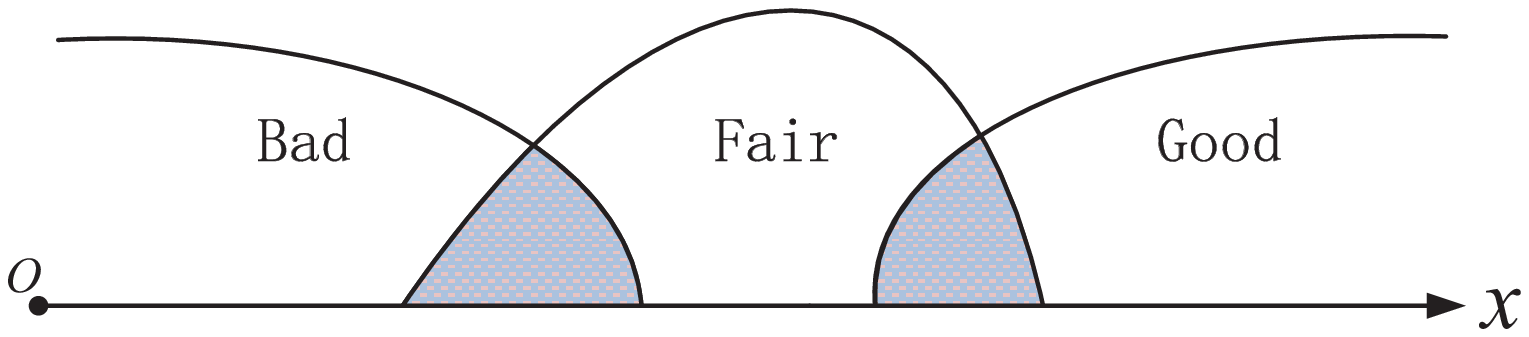}\label{DNumbersElements}}
     \hspace{0.1in}
  \caption{The comparison between the frame of discernment in Dempster-Shafer evidence theory and the
  problem domain in D numbers}\label{DSandDcomparison}
\end{figure}

Suppose there exists a task to assess a project. In the frame of Dempster-Shafer theory, the frame of discernment is $\{Good, Fair, Bad \}$, shown in Figure \ref{FrameDS}. A BPA could be constructed to express the expert's assessment:
\begin{equation}
\begin{array}{l}
 m(\{ Good \} ) = 0.2 \\
 m(\{ Fair \} ) = 0.7 \\
 m(\{ Fair, Bad \} ) = 0.1 \\
 \end{array}
\end{equation}

If another expert gives his assessment by using D numbers, the problem domain can be shown as Figure \ref{DNumbersElements}. The assessment is as follows.
\begin{equation}
\begin{array}{l}
 D(\{ Good \} ) = 0.2 \\
 D(\{ Fair \} ) = 0.6 \\
 D(\{ Fair, Bad \} ) = 0.1 \\
 \end{array}
\end{equation}

Note that the set of $\{Good, Fair, Bad \}$ in D numbers is not a frame of discernment, because the elements are not
mutually exclusive. In addition, the additive constraint is released in D numbers. In this example $ D(\{ Good \} ) + D(\{ Fair \} ) + D(\{ Fair, Bad \} ) = 0.9$. If $\sum\limits_{B \subseteq \Omega } {D(B) = 1}$, the information is said to be complete; If $\sum\limits_{B \subseteq \Omega } {D(B) < 1}$, the information is said to be incomplete.

If a problem domain is $\Omega = \{b_1, b_2, \cdots, b_i, \cdots, b_n\}$, where $b_i \in R$ and $b_i \ne b_j$ if $i \ne j$, a special form of D numbers can be expressed by
\begin{equation}
\begin{array}{l}
 D(\{ b_1 \} ) = v_1  \\
 D(\{ b_2 \} ) = v_2  \\
 \cdots \qquad \cdots \\
 D(\{ b_i \} ) = v_i  \\
 \cdots \qquad \cdots \\
 D(\{ b_n \} ) = v_n  \\
 \end{array}
\end{equation}
simple noted for $D = \{(b_1, v_1), (b_2, v_2), \cdots, (b_i, v_i), \cdots, (b_n, v_n) \}$, where $v_i > 0$ and  $\sum\limits_{i = 1}^n {v_i } \le 1$. Some properties of D numbers are introduced as follows.
\begin{property}
Permutation invariability. If there are two D numbers that
\[
D_1 = \{(b_1, v_1), \cdots, (b_i, v_i), \cdots, (b_n, v_n) \}
\]
and
\[
D_2 = \{ (b_n, v_n), \cdots, (b_i, v_i), \cdots, (b_1, v_1)\},
\]
then $D_1 \Leftrightarrow D_2$.
\end{property}

\begin{property}
Let $D = \{(b_1, v_1), (b_2, v_2), \cdots, (b_i, v_i), \cdots, (b_n, v_n) \}$ be a D numbers, the integration representation of $D$ is defined as
\begin{equation}\label{D_integration}
I(D) = \sum\limits_{i = 1}^n {b_i v_i }
\end{equation}
where $b_i \in R$, $v_i > 0$ and $\sum\limits_{i = 1}^n {v_i } \le
1$.
\end{property}

\subsection{Combination rule of D numbers}

\subsubsection{Relative matrix.} 
\begin{figure}[htbp]
\begin{center}
\centerline{
\includegraphics[width=10cm]{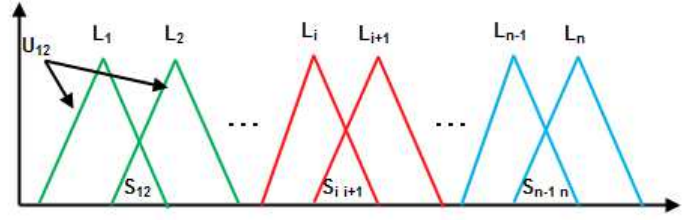}
}
\caption{Four linguistic constants.
}\label{linguistic example}
%(d) the US power-grid. (f)network of e-mail interchanges.
\vspace{0.2in}
\end{center}
\end{figure}

$n$ linguistic constants expressed in normal triangular fuzzy numbers are illustrated in Fig. \ref{linguistic example}. The area of intersection $S_{ij}$ and union $U_{ij}$  between any two triangular fuzzy numbers $L_i$ and $L_j$ can be can calculated to represent the non-exclusive degree between two D numbers. For example, the intersection $S_{12}$ and the union $U_{12}$ in Fig. \ref{linguistic example}. The non-exclusive degree $D_{ij}$ can be calculated as follows:
\begin{equation}\label{exclusive degree}
  D_{ij}=\frac{S_{ij}}{U_{ij}}
\end{equation}

 It should be emphasized that how to determine the non-exclusive degree depends on the application type. Due to the characteristic of the fuzzy numbers, we choose the area of intersection and union between two fuzzy numbers. A relative Matrix for these elements based on the non-exclusive degree can be build as below:
\begin{equation}\label{relative matrix example}
  R=
\left[
\begin{array}{ccccccc}
      &  L_1     &  L_2    & \ldots  &  L_i    &\ldots & L_n\\
 L_1  & 1        &  D_{12} &\ldots  &  D_{1i} &\ldots  &  D_{1n}  \\
 L_2  &  D_{21}  &  1      &\ldots  &  D_{2i} &\ldots  &  D_{2n}  \\
 \vdots  & \vdots  & \vdots &\ldots      &\vdots     &\ldots&  \vdots  \\
 L_i  &  D_{i1}  &  D_{i2} &\ldots  &  1 &\ldots &  D_{in}\\
 \vdots  & \vdots  & \vdots &\ldots      &\vdots     &\ldots&  \vdots  \\
 L_n  &  D_{n1}  &  D_{n2} &\ldots  &  D_{ni} &\ldots & 1
\end{array}
\right]
\end{equation}

%\begin{equation}\label{relative matrix example}
%  R=
%\left[
%\begin{array}{cccccc}
%      &  L_1   &  L_2 &...  &  L_i   &  L_{i+1}  \\
% L_1  &  D_{11}  &  D_{12}  &  D_{13}  &  D_{14}  \\
% L_2  &  D_{21}  &  D_{22}  &  D_{23}  &  D_{24}  \\
% L_1  &  D_{31}  &  D_{32}  &  D_{33}  &  D_{34}  \\
% L_1  &  D_{41}  &  D_{42}  &  D_{43}  &  D_{44}
%\end{array}
%\right]
%\end{equation}

\subsubsection{Exclusive coefficient.} The exclusive coefficient $\varepsilon$ is used to characterize the exclusive degree of the propositions in a assessment situation, which is got by calculating the average non-exclusive degree of these elements using the upper triangular of the relative matrix. Namely:
\begin{equation}\label{ce}
  \varepsilon=\frac{\sum_{i,j=1,i\neq{j}}^{n}{D_{ij}}}{n(n-1)/2}
\end{equation}
where $n$ is the number of the propositions in the assessment situation. Smaller the $\varepsilon$ is, the more exclusive the propositions of the application are. When $\varepsilon=0$, the propositions of application are completely mutually exclusive. That is, this situation is up to the requirements of the Dempster-Shafer evidence theory.

\textbf{The combination rule of D numbers.}  Firstly, the given D numbers should be discounted by the exclusive coefficient $\varepsilon$, which can guarantee the elements in the frame of discernment $\Omega$ to be exclusive. The D numbers can be discounted as below:
      \begin{eqnarray*}
        D(A_i)_\varepsilon=D(A_i).(1-\varepsilon)
      \end{eqnarray*}
      \begin{equation}\label{discount}
        D(\Theta)_\varepsilon=D(\Theta).(1-\varepsilon)+\varepsilon
      \end{equation}
      where $A_i$ is the elements in $\Omega$.

Then the combination rule of D numbers based on the exclusive coefficient is illustrated as follows.
\begin{equation}\label{conbination}
  D(A)_{\varepsilon}=\frac{\sum_{B\cap{C}=A}{D_1(B)_{\varepsilon}D_2(C)_{\varepsilon}}}{1-k}
\end{equation}
with
\begin{equation}\label{conflict}
  k=\sum_{B\cap{C}={\phi}}{D_1(B)_{\varepsilon}D_2(C)_{\varepsilon}}
\end{equation}
where $k$ is a normalization constant, called conflict because it measures the degree of conflict between $D_1$ and $D_2$.

One should note that, if $\varepsilon=0$, i.e, the elements in the frame of discernment $\Omega$ are completely mutually exclusive, the D numbers will not be discount by the exclusive coefficient. That is, the mutually exclusive situation of D numbers is completely the same with the Dempster-Shafer evidence theory.

\subsection{Decision making based on linguistic variable transformation}

In this section, we propose a so called linguistic variable transformation. If the hypothesis is exclusive with each other, the LVT is degenerated to PPT in transferable belief model.
\begin{definition}
The transformation is determined by the area ratio of the intersection $S_{ij}$ and the corresponding linguistic variable area.
\begin{equation}\label{LVT_A}
{D_{LVT}}(A) = D(A) + D(A,B)\frac{{\frac{{{S_{AB}}}}{{{S_A}}}}}{{\frac{{{S_{AB}}}}{{{S_A}}} + \frac{{{S_{AB}}}}{{{S_B}}}}}
\end{equation}
\begin{equation}\label{LVT_B}
{D_{LVT}}(B) = D(B) + D(A,B)\frac{{\frac{{{S_{AB}}}}{{{S_B}}}}}{{\frac{{{S_{AB}}}}{{{S_A}}} + \frac{{{S_{AB}}}}{{{S_B}}}}}
\end{equation}
where $S_{AB}$ denotes the intersection area between linguistic variable $A$ and linguistic variable $B$. $S_A$ and $S_B$ represents the area of linguistic variable $A$ and linguistic variable $B$, respectively.
\end{definition}

Without loss of generality, if the hypothesis is exclusive with each other, which means the intersection $S_{ij}=0$ between each linguistic variable, the discount ratio is $1:1$ in transferable belief model. It is worth noting that LVT is applicative when and only when one linguistic variable do not completely belongs to another linguistic variable. In the following part, a simple example is given to show how LVT works.
Suppose the original D number is
\[\begin{array}{l}
D(VP,P) = 0.8\\
D(P) = 0.2
\end{array}\]

\begin{figure}[htbp]
\begin{center}
\centerline{
\includegraphics[scale=0.8]{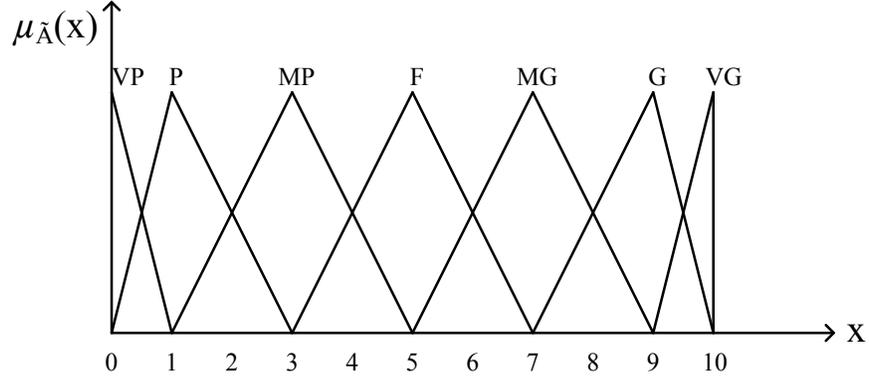}
}
\caption{The linguistic variable as described in Table \ref{VP}.
}\label{Linguistic_variables}
%(d) the US power-grid. (f)network of e-mail interchanges.
\vspace{0.2in}
\end{center}
\end{figure}

Fig. \ref{Linguistic_variables} shows the linguistic variable as described in Table \ref{VP}.

The areas can be obtained:
\[\begin{array}{l}
{S_{VP,P}} = 0.5 \times 1 \times 0.5 = 0.25\\
{S_{VP}} = 0.5 \times 1 \times 1 = 1\\
{S_P} = 0.5 \times 3 \times 1 = 1.5
\end{array}\]
Using Eqs.(\ref{LVT_A}) and (\ref{LVT_B}), the final D number by LVT is calculated
\[\begin{array}{l}
{D_{LVT}}(VP) = 0.8 \times \frac{{\frac{{0.25}}{1}}}{{\frac{{0.25}}{1} + \frac{{0.25}}{{1.5}}}} = 0.6\\
{D_{LVT}}(P) = 0.2 + 0.8 \times \frac{{\frac{{0.25}}{{1.5}}}}{{\frac{{0.25}}{1} + \frac{{0.25}}{{1.5}}}} = 0.4
\end{array}\]

\section{Application}
To demonstrate the efficiency and practicability of the proposed method, the example illustrated in \cite{chen2000extensions} is used. Suppose that a company desires to hire an engineer. After preliminary screening, three candidates $A_1$, $A_2$ and $A_3$ remain for further evaluation. A committee of three experts, $D_1$, $D_2$ and $D_3$ has been formed to conduct the interview and select the most suitable candidate. Five benefit criteria are considered: \\
(1) emotional steadiness ($C_1$) \\
(2) oral communication skill ($C_2$) \\
(3) personality ($C_3$) \\
(4) past experience ($C_4$) \\
(5) self-confidence ($C_5$)

\begin{figure}
\centering
\includegraphics[scale=0.6]{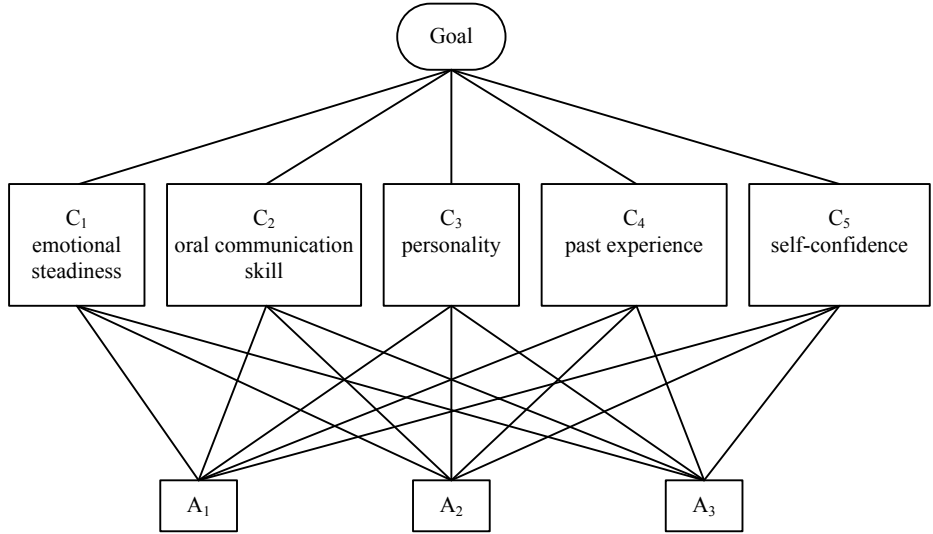}
\caption{The hierarchical structure} \label{A1_A2_A3}
\end{figure}

Fig. \ref{A1_A2_A3} shows the hierarchical structure of the decision process. The proposed method is now applied to solve this decision problem. Then, computational procedure is summarized as follows:

\emph{Step} 1: The experts use linguistic weighting variables to assess the importance of the criteria, which is shown in Table \ref{weight}.

\emph{Step} 2: Using the linguistic rating variables, the ratings of the three candidates by experts under all criteria is obtained as shown in Table \ref{ratings_by_experts}.

\emph{Step} 3: Convert the linguistic weighted evaluation shown in Tables \ref{weight} and \ref{ratings_by_experts} into triangular fuzzy numbers as shown in Table \ref{TFN}.

\emph{Step} 4: Calculating the ratio of the intersection between the obtained triangular fuzzy number area and the single linguistic variable area to construct the D numbers in each candidate with five criteria as shown in Table \ref{D_number_of_candidates}.

Note that the non-intersecting area represents that this part may belongs to arbitrary linguistic variables, where this ratio is defined as $D(\Theta )$.

\emph{Step} 5: The discounted D numbers with exclusive coefficient are aggregated by the combination rule of D numbers. The results are shown in Table \ref{combination_of D}.

\emph{Step} 6: The proposed linguistic variables transformation (LVT), converted a belief function to a probability function benefited to making a decision, is used to get the final combined D number of three candidates with five criteria as shown in Table \ref{D_number_LVT}.

\emph{Step} 7: The decision of choosing which one candidate is determined by the maximum supported degree of ``$Medium~Good$'' $MSD(MG)$. Calculate $MSD(MG)$ by the sum of ``$MG$'', ``$G$'', and ``$VG$'' of each candidate as
\[MS{D_{{A_1}}}(MG) = {D_{{A_1}}}(\{ MG\} ) + {D_{{A_1}}}(\{ G\} ) + {D_{{A_1}}}(\{ VG\} ) = 0.9444\]
\[MS{D_{{A_2}}}(MG) = {D_{{A_2}}}(\{ MG\} ) + {D_{{A_2}}}(\{ G\} ) + {D_{{A_2}}}(\{ VG\} ) = 0.9858\]
\[MS{D_{{A_3}}}(MG) = {D_{{A_3}}}(\{ MG\} ) + {D_{{A_3}}}(\{ G\} ) + {D_{{A_3}}}(\{ VG\} ) = 0.9828\]

\emph{Step} 8: According to the maximum supported degree of ``$Medium~Good$'', the ranking order of the three candidates is ${A_2} \succ {A_3} \succ {A_1}$, where ``$\succ$'' represents ``$better~than$''. Obviously, the best selection is candidate $A_2$.

\begin{table}[!h]
{\footnotesize%
\caption{The importance weight of the criteria}\label{weight}
\begin{center}
\begin{tabular}{p{2cm}p{2cm}p{2cm}p{1cm}}
\toprule
      & $E_1$ & $E_2$ & $E_3$  \\
\midrule
$C_1$ & H  & VH & MH  \\
$C_2$ & VH & VH & VH  \\
$C_3$ & VH & H  & H   \\
$C_4$ & VH & VH & VH  \\
$C_5$ & M  & MH & MH  \\
\bottomrule
\end{tabular}
\end{center}}
\end{table}

\begin{table}[!h]
{\footnotesize%
\caption{The ratings of the three candidates by experts under all criteria}\label{ratings_by_experts}
\begin{center}
\begin{tabular}{p{2cm}p{2cm}p{2cm}p{2cm}p{1cm}}
\toprule
   \multicolumn{1}{l}{Criteria}&\multicolumn{1}{l}{Candidates}&\multicolumn{3}{l}{Experts} \\
   \cmidrule(r){3-5}
     &    & $E_1$ & $E_2$ & $E_3$ \\
\midrule
  $C_1$  & $A_1$ & MG & G  & MG     \\
         & $A_2$ & G  & G  & MG     \\
         & $A_3$ & VG & G  & F      \\
  $C_2$  & $A_1$ & G  & MG & F      \\
         & $A_2$ & VG & VG & VG     \\
         & $A_3$ & MG & G  & VG     \\
  $C_3$  & $A_1$ & F  & G  & G      \\
         & $A_2$ & VG & VG & G      \\
         & $A_3$ & G  & MG & VG     \\
  $C_4$  & $A_1$ & VG & G  & VG     \\
         & $A_2$ & VG & VG & VG     \\
         & $A_3$ & G  & VG & MG     \\
  $C_5$  & $A_1$ & F  & F  & F      \\
         & $A_2$ & VG & MG & G      \\
         & $A_3$ & G  & G  & MG     \\
\bottomrule
\end{tabular}
\end{center}}
\end{table}

\begin{table}[!h]
{\footnotesize%
\caption{The triangular fuzzy numbers of three candidates with five criteria}\label{TFN}
\begin{center}
\begin{tabular}{p{0.15cm}p{2.2cm}p{2.2cm}p{2.2cm}p{2.2cm}p{2.2cm}}
\toprule
           & $C_1$ & $C_2$ & $C_3$ & $C_4$ & $C_5$  \\
\midrule
     $A_1$ & (4.87,6.56,7.92) & (4.83,6.77,8.38) & (5.01,6.81,8.03) & (7.22,8.38,8.67) & (1.90,3.17,4.43)    \\
     $A_2$ & (5.44,7.13,8.21) & (8.70,9.67,9.67) & (7.52,8.71,9.00) & (7.80,8.67,8.67) & (4.30,5.40,6.10)    \\
     $A_3$ & (5.56,6.96,7.44) & (6.77,8.38,9.34) & (6.30,7.81,8.71) & (6.07,7.51,8.38) & (3.97,5.23,6.10)    \\
\bottomrule
\end{tabular}
\end{center}}
\end{table}

\begin{table}[!h]
{\footnotesize%
\caption{The obtained D number of three candidates with five criteria}\label{D_number_of_candidates}
\begin{tabular*}{\columnwidth}{@{\extracolsep{\fill}}@{~~}lrrr@{~~}}
\toprule
 $C_i$ & $A_1$ & $A_2$ & $A_3$  \\
\midrule
$C_1$ & $D(\{MP\},\{F\})=0.0015$ & $D(\{F\},\{MG\})=0.2381$ & $D(\{F\},\{MG\})=0.3244$  \\
      & $D(\{F\})=0.0744$        & $D(\{MG\})=0.5853$       & $D(\{MG\})=0.6330$        \\
      & $D(\{F\},\{MG\})=0.3279$ & $D(\{MG\},\{G\})=0.1716$ & $D(\{MG\},\{G\})=0.0415$  \\
      & $D(\{MG\})=0.4215$       & $D(\Theta)=0.0050$       & $D(\Theta)=0.0011$        \\
      & $D(\{MG\},\{G\})=0.0826$ &                          &                           \\
      & $D(\Theta)=0.0921$       &                          &                           \\
$C_1$ & $D(\{MP\},\{F\})=0.0021$ & $D(\{MG\},\{G\})=0.0312$ & $D(\{F\},\{MG\})=0.0057$  \\
      & $D(\{F\})=0.0529$        & $D(\{G\})=0.3377$        & $D(\{MG\})=0.1412$        \\
      & $D(\{F\},\{MG\})=0.2817$ & $D(\{G\},\{VG\})=0.4032$ & $D(\{MG\},\{G\})=0.3891$  \\
      & $D(\{MG\})=0.4611$       & $D(\{VG\})=0.0596$       & $D(\{G\})=0.3077$         \\
      & $D(\{MG\},\{G\})=0.1486$ & $D(\Theta)=0.1983$       & $D(\{G\},\{VG\})=0.0230$  \\
      & $D(\Theta)=0.0536$       &                          & $D(\Theta)=0.1333$        \\
$C_1$ & $D(\{F\})=0.0141$        & $D(\{MG\})=0.4639$       & $D(\{F\},\{MG\})=0.0579$  \\
      & $D(\{F\},\{MG\})=0.3310$ & $D(\{MG\},\{G\})=0.4907$ & $D(\{MG\})=0.3889$        \\
      & $D(\{MG\})=0.5039$       & $D(\Theta)=0.0454$       & $D(\{MG\},\{G\})=0.3832$  \\
      & $D(\{MG\},\{G\})=0.1091$ &                          & $D(\{G\})=0.0352$         \\
      & $D(\Theta)=0.0419$       &                          & $D(\Theta)=0.1348$        \\
$C_1$ & $D(\{MG\})=0.0416$       & $D(\{MG\})=0.5141$       & $D(\{F\},\{MG\})=0.1088$  \\
      & $D(\{MG\},\{G\})=0.6060$ & $D(\{MG\},\{G\})=0.4377$ & $D(\{MG\})=0.5370$        \\
      & $D(\{G\})=0.1942$        & $D(\Theta)=0.0482$       & $D(\{MG\},\{G\})=0.2873$  \\
      & $D(\Theta)=0.1582$       &                          & $D(\Theta)=0.0669$        \\
$C_1$ & $D(\{P\},\{MP\})=0.1463$ & $D(\{MP\},\{F\})=0.0878$ & $D(\{MP\},\{F\})=0.1528$  \\
      & $D(\{MP\})=0.5939$       & $D(\{F\})=0.6235$        & $D(\{F\})=0.6349$         \\
      & $D(\{MP\},\{F\})=0.2479$ & $D(\{F\},\{MG\})=0.2490$ & $D(\{F\},\{MG\})=0.1979$  \\
      & $D(\Theta)=0.0119$       & $D(\Theta)=0.0397$       & $D(\Theta)=0.0144$        \\
\bottomrule
\end{tabular*}
}
\end{table}

\begin{table}[!h]
{\footnotesize%
\caption{The combination of discounted D number}\label{combination_of D}
\begin{tabular*}{\columnwidth}{@{\extracolsep{\fill}}@{~~}rrr@{~~}}
\toprule
          $A_1$ & $A_2$ & $A_3$  \\
\midrule
          $D(\{P\},\{MP\})=0.00083$ & $D(\{MP\},\{F\})=0.00027$ & $D(\{MP\},\{F\})=0.00016$     \\
          $D(\{MP\})=0.00346$       & $D(\{F\})=0.01096$        & $D(\{F\})=0.01418$            \\
          $D(\{MP\},\{F\})=0.00146$ & $D(\{F\},\{MG\})=0.00561$ & $D(\{F\},\{MG\})=0.00569$     \\
          $D(\{F\})=0.03973$        & $D(\{MG\})=0.54548$       & $D(\{MG\})=0.97317$           \\
          $D(\{F\},\{MG\})=0.01988$ & $D(\{MG\},\{G\})=0.05219$ & $D(\{MG\},\{G\})=0.00408$     \\
          $D(\{MG\})=0.92027$       & $D(\{G\})=0.38467$        & $D(\{G\})=0.00265$            \\
          $D(\{MG\},\{G\})=0.01127$ & $D(\{G\},\{VG\})=0.00049$ & $D(\{G\},\{VG\})=0.00001$     \\
          $D(\{G\})=0.00278$        & $D(\{VG\})=0.00007$       & $D(\Theta)=0.00006$           \\
          $D(\Theta)=0.00032$       & $D(\Theta)=0.00026$       &                               \\
\bottomrule
\end{tabular*}
}
\end{table}

%\begin{table}[htbp]
%\caption{Combination}\label{combination}
%\resizebox{\textwidth}{!}{
%\begin{tabular}{rrr}
%\toprule
%          $A_1$ & $A_2$ & $A_3$  \\
%\midrule
%          $D(\{P\},\{MP\})=0.00083$ & $D(\{MP\},\{F\})=0.00027$ & $D(\{MP\},\{F\})=0.00016$     \\
%          $D(\{MP\})=0.00346$       & $D(\{F\})=0.01096$        & $D(\{F\})=0.01418$            \\
%          $D(\{MP\},\{F\})=0.00146$ & $D(\{F\},\{MG\})=0.00561$ & $D(\{F\},\{MG\})=0.00569$     \\
%          $D(\{F\})=0.03973$        & $D(\{MG\})=0.54548$       & $D(\{MG\})=0.97317$           \\
%          $D(\{F\},\{MG\})=0.01988$ & $D(\{MG\},\{G\})=0.05219$ & $D(\{MG\},\{G\})=0.00408$     \\
%          $D(\{MG\})=0.92027$       & $D(\{G\})=0.38467$        & $D(\{G\})=0.00265$            \\
%          $D(\{MG\},\{G\})=0.01127$ & $D(\{G\},\{VG\})=0.00049$ & $D(\{G\},\{VG\})=0.00001$     \\
%          $D(\{G\})=0.00278$        & $D(\{VG\})=0.00007$       & $D(\Theta)=0.00006$           \\
%          $D(\Theta)=0.00032$       & $D(\Theta)=0.00026$       &                               \\
%\bottomrule
%\end{tabular}}
%\end{table}

\begin{table}[!h]
{\footnotesize%
\caption{The final D number by linguistic variables transformation}\label{D_number_LVT}
\begin{tabular*}{\columnwidth}{@{\extracolsep{\fill}}@{~~}rrr@{~~}}
\toprule
          $A_1$ & $A_2$ & $A_3$  \\
\midrule
          $D(\{VP\})=0.000094$ & $D(\{VP\})=0.000075$ & $D(\{VP\})=0.000018$     \\
          $D(\{P\})=0.000503$  & $D(\{P\})=0.000025$  & $D(\{P\})=0.000006$      \\
          $D(\{MP\})=0.004567$ & $D(\{MP\})=0.000153$ & $D(\{MP\})=0.000084$     \\
          $D(\{F\})=0.050428$  & $D(\{F\})=0.013921$  & $D(\{F\})=0.017102$      \\
          $D(\{MG\})=0.935062$ & $D(\{MG\})=0.570672$ & $D(\{MG\})=0.977771$     \\
          $D(\{G\})=0.009252$  & $D(\{G\})=0.414680$  & $D(\{G\})=0.004996$      \\
          $D(\{VG\})=0.000094$ & $D(\{VG\})=0.000474$ & $D(\{VG\})=0.000023$     \\
\bottomrule
\end{tabular*}
}
\end{table}
\section{Conclusion}
One of the assumptions to apply the Dempster-Shafer evidence evidence theory is that all the elements in the frame of discernment should be mutually exclusive. However, it is difficult to meet the requirement in the real-world applications. In this paper, a new mathematic tool to model uncertain information, called as D numbers, is used to model and combine the domain experts' opinions under the condition that the linguistic constants are not exclusive with each other. An exclusive coefficient is proposed to discount the D numbers. After the discounted D numbers are obtained, the domain experts' opinion can be fused based on our proposed combination rule of D numbers. It is inevitable to handle the experts' subjective opinion, the proposed D numbers is promising methodology since it provides more flexible way than classical evidence theory to deal with uncertain. Though the D numbers theory is illustrated to handle linguistic decision making problems in this paper, the proposed theory can handle other situation when the hypothesis is not exclusive with each other. Fianlly, when the elements in the frame of discernment are mutually exclusive, D numbers theory is degenerated as classical Dempster Shafer evidence theory.

%\cite{du2014new,li2013new,Linyan2014Developing,Li2014Modeling,Li2014vulnerability,Xi2014Bargaining,Chenwei2014new}
% One of the hypothesis of the Dempster-Shafer theory is that elements in the frame of discernment should be mutually exclusive, but it is difficult to meet the real-world applications. The risk of contaminant intrusion is a function of three elements -- a pathway, a driving force, and a contamination source. And the exclusive degree of the these elements is correlated to the results of the risk estimating. In this paper, we proposed a new model to estimating the risk of the contamination intrusion of the water distribution network. The proposed D number can guarantee that the non-exclusive elements can hold a property of mutually exclusive. And the result show that the proposed method can provide a reasonable effect.

\section*{Acknowledgements}
The author greatly appreciates Professor Shan Zhong, the China academician of the Academy of Engineering, for his encouragement to do this research. The author also greatly appreciates Professor Yugeng Xi in Shanghai Jiao Tong University for his support to this work. Professor Sankaran Mahadevan in Vanderbilt University discussed many relative topics about this work. The author's Ph.D students in Shanghai Jiao Tong University, Peida Xu and Xiaoyan Su, Ph.D students in Southwest University, Xinyang Deng and Daijun Wei, the graduate students in Southwest University Yajuan Zhang, Bingyi Kang, Xiaoge Zhang, Shiyu Chen, Yuxian Du, Cai Gao, have discussed the topic of D numbers. The undergraduate student Li Gou does some editorial work for this paper. This work is partially supported by National Natural Science Foundation of China, Grant No. 61174022, Chongqing Natural Science Foundation, Grant No. CSCT, 2010BA2003, Program for New Century Excellent Talents in University, Grant No.NCET-08-0345, Beihang University (Grant No.BUAA-VR-14KF-02).


%% References
%% Usage of  \cite is as follows:
%%    \cite{key}          ==>>  [#]
%%    \cite[chap. 2]{key} ==>>  [#, chap. 2]

\section*{References}
\bibliographystyle{elsarticle-num-names}
\bibliography{myref}






% \end{thebibliography}

\end{document}